\documentclass[journal,twoside,web]{ieeecolor}
\usepackage{generic}
\usepackage{cite}
\usepackage{amsmath,amssymb,amsfonts}
\usepackage{algorithmic}
\usepackage{graphicx}
\usepackage{textcomp}
\usepackage{subfigure}
\usepackage{stfloats}
\usepackage{multirow}
\usepackage{times}
\usepackage{soul}
\usepackage{url}
\usepackage{booktabs}
\usepackage{utfsym}
\usepackage[ruled,linesnumbered]{algorithm2e}
\usepackage{bbold}
\usepackage{bbding}
\usepackage{hyperref}
\usepackage{lineno}
\def\BibTeX{{\rm B\kern-.05em{\sc i\kern-.025em b}\kern-.08emT\kern-.1667em\lower.7ex\hbox{E}\kern-.125emX}}
\begin{document}

\title{MedFILIP: Medical Fine-grained Language-Image Pre-training}
\author{Xinjie Liang, Xiangyu Li, \IEEEmembership{Member, IEEE}, Fanding Li, Jie Jiang, Qing Dong, Wei Wang, \IEEEmembership{Member, IEEE}, Kuanquan Wang, \IEEEmembership{Senior Member, IEEE}, Suyu Dong, \IEEEmembership{Member, IEEE}, Gongning Luo, \IEEEmembership{Member, IEEE}, and Shuo Li,\IEEEmembership{Senior Member, IEEE}
  \thanks{Manuscript received ***; revised ***; accepted ***. Date of publication *** ***, ***; date of current version *** **, ****. This work was supported by the Key Research \& Development Program of Heilongjiang Province under Grant 2023X01A08, the National Natural Science Foundation of China under Grants 62272135, 62372135; the Postdoctoral Fellowship Program of CPSF under Grant Number GZC20242214; the China Postdoctoral Science Foundation under Grants 2024M754207. \emph{(Corresponding author: Xiangyu Li; Gongning Luo.)}}
  \thanks{Xinjie Liang, Xiangyu Li (e-mail: lixiangyu@hit.edu.cn), Fanding Li, Jie Jiang, and Kuanquan Wang are with the School of Computer Science and Technology, Harbin Institute of Technology, Harbin, China.
    Qing Dong is with the Department of Thoracic Surgery at No. 4 Affiliated Hospital, Harbin Medical University, Harbin, China.
    Wei Wang is with the School of Computer Science and Technology, Harbin Institute of Technology, Shenzhen, China.
    Suyu Dong is with the College of computer and control engineering, Northeast Forestry University, Harbin, China.
    Gongning Luo (gongning.luo@kaust.edu.sa) is with the Computer, Electrical and Mathematical Sciences \& Engineering Division, King Abdullah University of Science and Technology, Thuwal, Saudi Arabia.
    Shuo Li (e-mail: shuo.li11@case.edu) is with the Department of Biomedical Engineering and Department of Computer and Data Science Case Western Reserve University, Cleveland, OH, USA.}}

\maketitle
\begin{abstract}
  Medical vision-language pretraining (VLP) that leverages naturally-paired medical image-report data is crucial for medical image analysis.
  However, existing methods struggle to accurately characterize associations between images and diseases, leading to inaccurate or incomplete diagnostic results.
  In this work, we propose MedFILIP, a fine-grained VLP model, introduces medical image-specific knowledge through contrastive learning, specifically:
  1) An information extractor based on a large language model is proposed to decouple comprehensive disease details from reports, which excels in extracting disease deals through flexible prompt engineering, thereby effectively reducing text complexity while retaining rich information at a tiny cost.
  2) A knowledge injector is proposed to construct relationships between categories and visual attributes, which help the model to make judgments based on image features, and fosters knowledge extrapolation to unfamiliar disease categories.
  3) A semantic similarity matrix based on fine-grained annotations is proposed, providing smoother, information-richer labels, thus allowing fine-grained image-text alignment.
  4) We validate MedFILIP on numerous datasets, e.g., RSNA-Pneumonia, NIH ChestX-ray14, VinBigData, and COVID-19. For single-label, multi-label, and fine-grained classification, our model achieves state-of-the-art performance, the classification accuracy has increased by a maximum of 6.69\%.
  The code is available in \url{https://github.com/PerceptionComputingLab/MedFILIP}.
\end{abstract}


\section{Introduction}
\label{sec:introduction}
\begin{figure}[t]
  \begin{minipage}[t]{\columnwidth}
    \centering
    \includegraphics[width=\columnwidth]{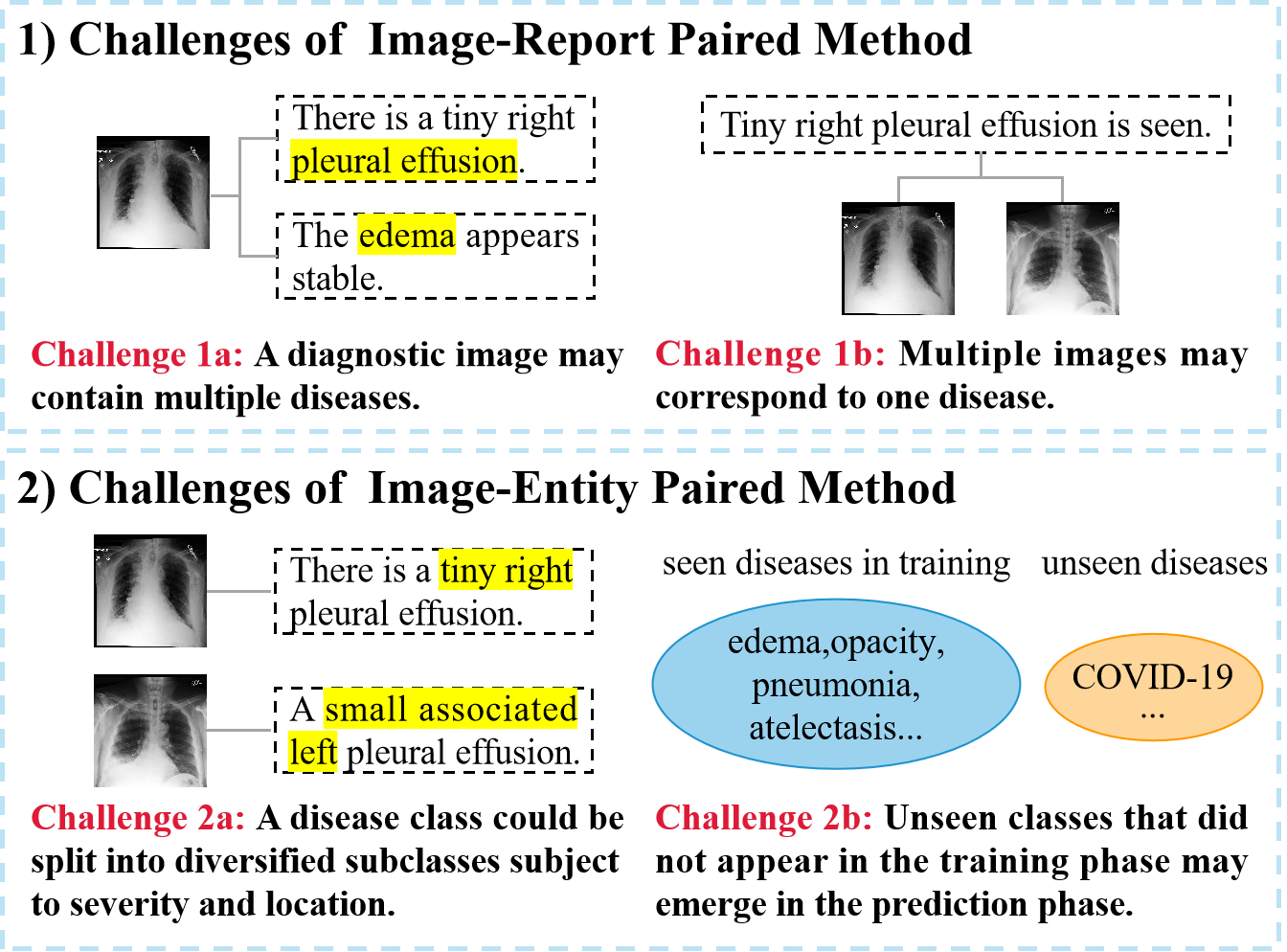}
    \caption{
      Existing medical VLP methods are limited in adequately characterizing the relationships between images and diseases.
      Using reports as supervision risks (1a) confusing disease descriptions from a single report and (1b) separating same-disease samples from different cases into distinct classes, leading to inaccurate classification. Using disease entities as supervision (2a) disregards subclasses, inhibiting customized care fitting a patient's circumstances, and (2b) can not classify classes unseen in the training dataset, lacking generalization.
    }
    \label{fig:0}
  \end{minipage}
\end{figure}

\begin{figure*}[h]
  \centering
  \includegraphics[width=\textwidth]{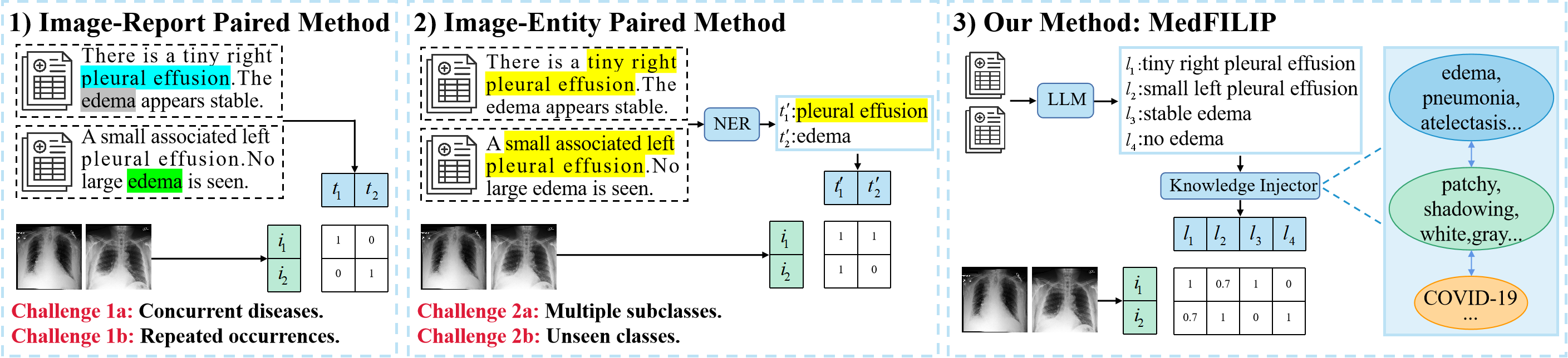}
  \caption{
    MedFILIP overcomes the limitations of previous methods by effectively modeling the complex relationship between images and reports.
    To address concurrent diseases and repeated occurrences, MedFILIP designs a GPT-IE to compress reports into fine-grained entities.
    To address overlooking subclasses, MedFILIP retains details like location and severity in the fine-grained entities.
    To improve the generalization of unseen categories, MedFILIP introduces an IKI to leverage image-specific knowledge from seen classes to guide inferences for unseen classes.
  }
  \label{fig:1}
\end{figure*}

\IEEEPARstart{M}{edical} vision-language pretraining (VLP) that leverages naturally-paired image-report data has proven to be a pivotal component for the medical image analysis tasks (MIA)~\cite{boecking2022making,li2024mlip,liu2023improving}.
Data scarcity, an ongoing challenge in MIA can be significantly mitigated by utilizing medical reports as supervision \cite{zhang2022convirt,wu2023medklip,wang2022medclip}, because reports occur alongside images naturally in clinical records, without requiring human labeling.
Additionally, reports are derived from real patient cases and inherently relate to the corresponding images, better reflecting real-world clinical scenarios and practice \cite{irvin2019chexpert,johnson2019mimic}.
Medical VLP methods that exploit this organically paired data excel in capturing the interplay between images and reports. This synergy is crucial for the effective modeling of image-disease relationships and essential for a variety of downstream medical applications.

However, existing medical VLP methods have limitations in adequately characterizing the relationships between images and diseases, primarily due to the challenge of harnessing the rich yet complex information contained within the diagnostic reports, as shown in Fig.~\ref{fig:0}.
There are two general medical VLP methods based on contrastive learning:
(1) \textbf{Image-report paired methods}, simply pair medical images with diagnostic reports for contrastive learning~\cite{radford2021clip,zhang2022convirt}.
However, using the entire report to match a medical image can lead to confusion when reports describe multiple diseases, as shown in Fig.~\ref{fig:1}.
Furthermore, treating images from different cases as distinct classes may result in incorrectly treating two samples with the same disease, but from different cases, as separate classes, as shown in Fig.~\ref{fig:1}.
(2) \textbf{Image-entity paired methods}, first extract disease category information from reports and then combine category information with prompt templates as the textual supervision for contrastive learning ~\cite{wang2022medclip}.
Although such methods effectively reduce the complexity of text, they result in the loss of rich contextual information in the reports.
Since a disease class could be split into diversified subclasses subject to severity and location, overlooking these details can lead to an inability to provide customized care plans suited to a patient's specific circumstances\cite{oakden2020hidden,dai2023efficient}.
Moreover, classes not encountered during the training phase may appear during the prediction phase.
Relying solely on categories without extra information will lead to a lack of generalization capability for these unseen categories.

To better capture the cross-modal correlations between images and texts, we propose MedFILIP, a medical vision-language pre-training method leveraging two types of fine-grained annotations.
The MedFILIP designs an information extractor based on a large language model (GPT-IE) to compress full diagnostic radiology reports into \textbf{fine-grained entities}. GPT-IE deals with concurrent diseases and repeated occurrences by decoupling disease information from lengthy reports and improves the model's understanding of complex intra-class variations by retaining critical information such as disease location and severity in the fine-grained entities.
The proposed MedFILIP introduces an image-specific knowledge Injector (IKI) that transforms each disease category into \textbf{fine-grained explanations} composed of visual attributes to enhance vision-language alignment. By employing attribute-level annotations, IKI captures the fine-grained relationships between diseases and visual attributes. This explains how models make judgments based on image features and enables the model to leverage the learned conceptual relationships from seen classes to guide inferences for unseen classes.
In order to fully utilize the \textbf{fine-grained entities and explanations}, we calculate a semantic similarity matrix (SSM) from them. SSM captures correspondences between images and texts more precisely, providing a more fine-grained supervisory signal that aids the model in learning to extract visual features and establish relationships between visual and textual features.

The key contributions of our method are as follows:

We present GPT-IE, a pioneering approach that leverages a large language model to extract triplet information—comprising disease severity, disease location, and disease category—from diagnostic reports. This decoupling process reduces textual complexity while preserving the rich information essential for medical vision-language pre-training.

Unlike previous approaches that do not fully utilize the wealth of medical knowledge available, IKI integrates this prior knowledge by establishing a mapping between disease categories and their visual-attribute descriptions. This strategy facilitates the transfer of knowledge from known to novel categories, thereby enabling zero-shot prediction for disease categories that have not been previously encountered.

To capture the similarity relationships between images and texts, we present SSM as an innovative alternative to the traditional identity matrix or multi-hot matrix labels employed in contrastive learning.
SSM is based on texts composed of extracted entities and image-specific explanations. The similarity between the texts is used as a proxy for the similarity between images and texts, ensuring that the vision-language alignments are rooted in precise, relevant labels. 

We introduce a new dataset comprising 141,171 images based on the MIMIC-CXR dataset~\cite{johnson2019mimic}. We enhance the texts with fine-grained entities and explanations, which reduce redundant information while retaining necessary disease details and introducing medical prior knowledge. This newly established large-scale medical multimodal dataset supports the training of MedFILIP and serves as a valuable resource for the broader research community, addressing the scarcity of fine-grained, annotated medical datasets.

\section{Related Work}

\subsection{Medical Vision-Language Contrastive Learning.}

To address the scarcity of annotated medical image-related data, numerous studies have employed text as supervisory information~\cite{boecking2022making,Flamingo,LLaVA}. The text contains rich and precise signals, and using data where images accompanied by text can reduce the need for additional manual labeling~\cite{LLaVA-Med,LLM_CXR,chatcad}. For example, CLIP~\cite{radford2021clip} collects data from the web to obtain a large number of image-text pairs. ConVIRT~\cite{zhang2022convirt} pairs diagnostic reports obtained during clinical diagnostic processes with corresponding images. MedCLIP~\cite{wang2022medclip} augments unpaired data by converting it into paired image-text data as a supplement. However, previous methods lack the ability to represent fine-grained visual features of the disease. Our approach improves the model's fine-grained representation ability by establishing a mapping between disease categories and their visual-attribute descriptions, thereby associating disease categories with their relevant attributes within the model. 

The medical image-text contrastive learning method employs a two-stream flow contrastive learning strategy to map images and text into a unified representation space, achieving semantic alignment between visual and textual modalities.
Image-report paired methods pair medical images with radiology reports for contrastive learning ~\cite{huang2021gloria,zhang2022convirt}; however, they encounter limitations in scenarios involving current diseases and repeated occurrences.
Image-entity paired methods apply named entity recognition (NER)~\cite{irvin2019chexpert} tools to extract disease categories from diagnostic texts, then combine category information with additional information as the textual supervision for contrastive learning~\cite{wang2022medclip,wu2023medklip}; however, they struggle to handle intra-class differences and generalize to novel classes.
Our approach decouples disease information from diagnostic reports, thus addressing the common problem of concurrent diseases and repeated occurrences.
And through the preservation of detailed information, our method offers an accurate understanding of various disease subclasses.

\begin{figure*}[h]
  \centering
  \includegraphics[width=\textwidth]{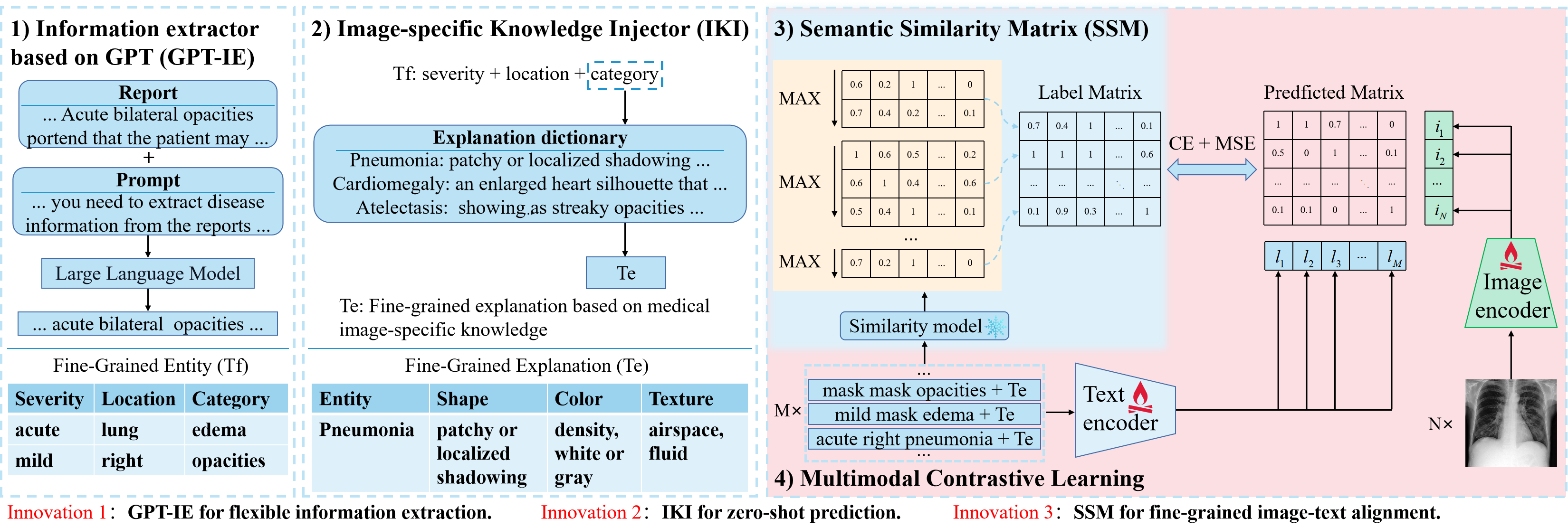}
  \caption{
    The proposed MedFILIP balances text complexity and information richness by distilling diagnostic reports into refined labeling formats and enhancing labels with medical image-specific knowledge.
    Specifically: 1) Diagnostic reports are processed by GPT-IE to obtain fine-grained entities.
    2) IKI is used to augment disease entities with medical image-specific knowledge.
    3) SSM is calculated based on the combination of fine-grained entities and explanations to establish clearer similarity-relationships between medical images and disease descriptions.
    4) Multimodal contrastive learning is conducted under the supervision of SSM.
    Additionally, we randomly mask the texts to address the problem of partially missing labels.}
  \label{fig:2}
\end{figure*}

\subsection{Fine-grained Representation Learning.}
Fine-grained representation learning refers to the process of learning detailed and discriminative representations for objects within a specific category by capturing the fine-grained relationship between image regions and textual attributes~\cite{FG1,FG3,FG4,FG5}.
Previous methods such as PLIP\cite{FG0} focused on the attribute-level distinctions between different categories, overlooking the connections that exist between categories.
Using additional medical knowledge can help the model implicitly establish relationships between medical entities\cite{Med_know,Med_know2}.
In our method, disease categories are not only transformed into combinations of image-specific attributes but are also enriched with additional medical knowledge, such as anatomical, pathological, and imaging correlates, which are missing in the traditional fine-grained representation learning approach.
This allows the model to leverage both more fine-grained visual features and prior knowledge from the medical domain, providing a more comprehensive characterization of the relationships between images and diseases.

\section{METHODOLOGY}

\subsection{GPT-IE for Fine-grained Label Extraction}
GPT-IE leverages the extensive knowledge encapsulated within large language model (LLM) to extract fine-grained entities from diagnostic reports.
This extraction process decouples disease information from lengthy reports, converting the one-to-one image-report relationship into a one-to-many relationship between image and diseases, which is instrumental in differentiating concurrent diseases and various instances of the same disease.
Additionally, GPT-IE captures detailed disease characteristics, such as location and severity, enabling the creation of a fine-grained medical dataset, which is pivotal for training deep learning models to handle intra-class variations, ultimately enhancing patient-specific care.

Distinct from traditional Named Entity Recognition (NER) systems~\cite{irvin2019chexpert}, which rely on predefined rules and lack flexibility, GPT-IE uses the LLM to extract entities without more human design.
The innovative use of prompt engineering enables LLM to perform specialized entity extraction without extensive setup~\cite{ner1_2022,ner2_2023}.
Designing suitable prompts can guide LLM to look for specific patterns, we employ prompts such as ``Each extracted piece of disease-related information should follow the format: severity of disease + location of disease + category of disease.”
Thus, GPT-IE can extract expected fine-grained entities, directly addressing the specific needs of medical data labeling. The extraction process is defined by:
\begin{align}
  \{severity, location, category\}= f_{IE}(report + prompt),
\end{align}%
where $f_{IE}$ is the information extractor, and the fine-grained entities consist of severity (e.g., acute, chronic, mild), location (e.g., left-sided, bilateral, mediastinal), and category (e.g., pneumonia, pneumothorax, pulmonary nodules).
Note that not all triples contain complete information; those missing severity or location will be preserved, while triples lacking the category will be removed.
For a balanced consideration of both cost and performance, our primary experiments were conducted using GPT-3.5-Turbo~\cite{ouyang2022instructgpt}.
The granular details of the extraction method, along with the LLM's prompt design strategies, are comprehensively elaborated in the supplementary material.

\textbf{Summarized advantages.}
GPT-IE introduces a cost-efficient framework capable of diverse information extraction needs.
By dynamically adjusting prompts based on actual needs, GPT-IE converts diagnostic reports expected into fine-grained disease information, which can reduce the complexity of supervisory text while retaining the disease details.

\subsection{IKI for Improving Generalization}
The proposed Image-specific Knowledge Injector (IKI) enables unseen category classification by correlating disease categories with visual features and prior knowledge.
Distinguishing between ``localized patchy shadowing or consolidative opacities" and ``darkened area between the lung and chest wall" is more straightforward than discerning between ``pneumonia" and ``pneumothorax"~\cite{li2022grounded,qin2022medical,chen2023knowledge}.
Building on this, IKI deconstructs abstract disease categories into these distinct, image-specific attributes. This decomposition enables the model to understand the relationships between visual features and the defining characteristics of clinical categories during the multimodal pretraining phase. Consequently, the model is armed with a conceptual framework that allows it to recognize features of known diseases and apply this understanding to identify unseen conditions, since visual attributes such as shape and color are common across disease categories.
Additionally, image-specific explanations will be expanded by medical knowledge to form a complete disease description. For example, the description of pneumonia is “An infection causing consolidation, often seen as patchy or confluent white areas within the lung fields”, where visual attributes and medical prior knowledge are both included. This allows the model to leverage both more fine-grained visual features and prior knowledge from the medical domain, providing more comprehensive relationships between images and diseases.

Different from the traditional image classification methods where an image $I$ is associated with a single label from a predefined set of classes, our approach considers a set of attributes $A=\{a_1,a_2,...,a_n\}$.
Each attribute $a_i$ represents a specific, identifiable feature in the image.
For a given $I$, a binary attribute vector $a_I=[a^I_1,a^I_2,...,a^I_n]$ can be defined, where $a^I_i$ is 1 if the attribute $a_i$ is present in the image and 0 otherwise. The classification basis is deconstructed as follows:
\begin{align}
  a_I=D(I),
\end{align}%
where $D$ is a deep neural network that deduces the attribute representation from the image.
With the extracted attribute vector $a_I$, the model predicts the disease by analyzing the correspondences between attributes and clinical classes.

The above method is operationalized through image-text contrastive learning.
Rather than explicitly calculating $a_I$, the model infers it by evaluating the similarity between image features and textual descriptions.
For instance, if there is a high similarity between an image and the text "localized areas of shadowing", it implies that the image possesses the attributes described in the text.
The model can then deduce the most likely disease category based on these attributes.

We map the category information from the fine-grained entity obtained by GPT-IE to an explanation based on a medical image-specific knowledge dictionary:
\begin{align}
  Te = Dict(Tf[categories]),
\end{align}%
where $Te$ is the explanation of disease, $Dict$ is a repository mapping text descriptions to each category, $Tf$ is the fine-grained entities. By using the category information in the fine-grained entities, relevant explanations of the disease can be obtained from the $Dict$.
More details of the $Dict$ can be found in the supplementary material.

\textbf{Summarized advantages.}
By transferring abstract disease categories into descriptive explanations encapsulating attributes like shape, structure, color, and texture, IKI helps the model to establish explicit connections between diseases and visual features, thereby allowing the model to recognize unseen diseases like COVID-19 based on the relationships learned between explanatory texts and manifested image patterns, without requiring direct training examples.
This process is more interpretable because it grounds the categorization in observable attributes rather than abstract categories.

\subsection{SSM for Finer-grained Image-text Alignment}
\begin{figure}[t]
  \begin{minipage}[t]{\columnwidth}
    \centering
    \includegraphics[width=\columnwidth]{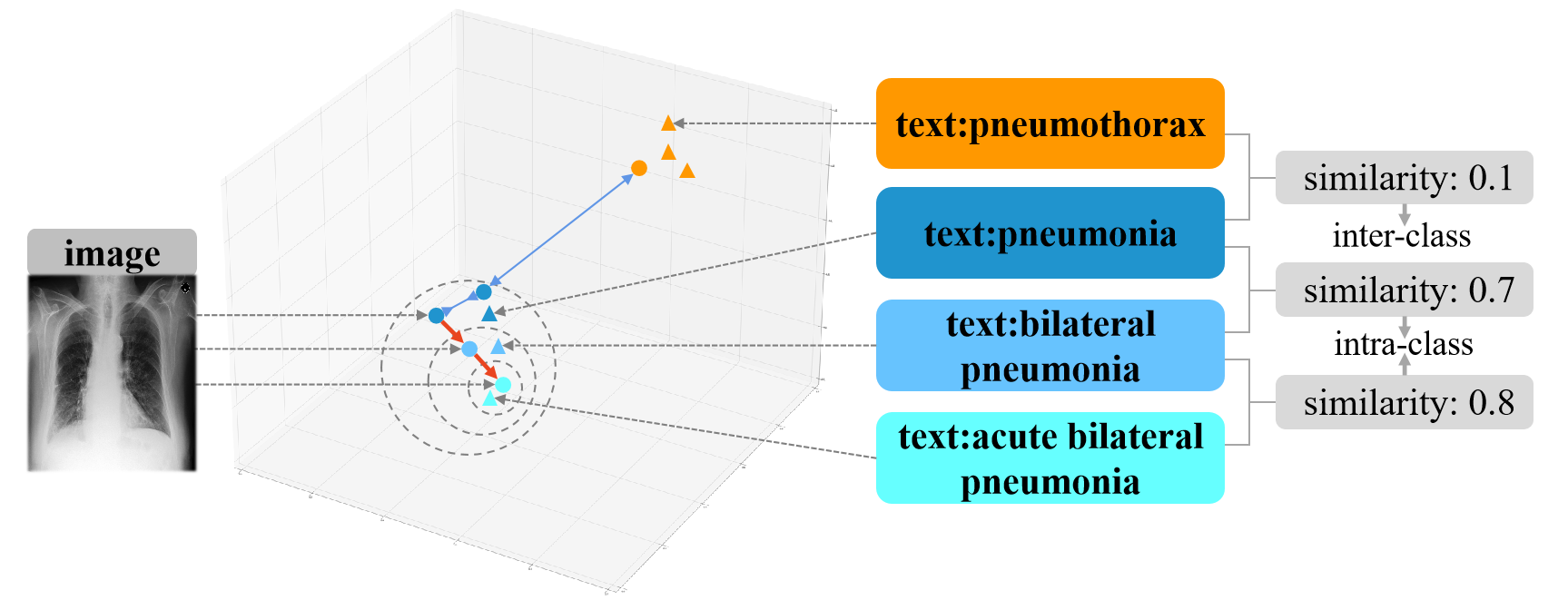}
    \caption{
      Our method maps image and text features into an embedding space, where it strategically clusters similar features while distancing dissimilar ones, and it also finely groups closely related subclasses based on semantic similarity.
    }
    \label{fig:4}
  \end{minipage}
\end{figure}

The proposed semantic similarity matrix (SSM) offers a more precise label representation that better captures the relationships between medical images and their textual annotations.
SSM is used to replace the multi-hot matrix for enhanced visual-language alignment, by using a continuum of values ranging from 0 to 1 instead of binary states (0 or 1), the SSM can not only realize clustering of different diseases but also realize internal clustering of different subclasses, as shown in Fig.~\ref{fig:4}.
Meanwhile, unlike previous methods where each image contains only one positive label for a specific condition\cite{huang2021gloria,wang2022medclip}, the SSM can handle situations where each image corresponds to one or more disease labels, thus enabling the model to perform multi-label classification.
We calculate the SSM based on structured labels that are formed by fine-grained entities and explanations.
The calculation process hinges on the premise that structured labels encapsulate the essential semantic essence of medical images; hence, these label interrelationships can act as a proxy for the correspondences between images and structured labels.
Specifically, the calculation process is as follows:

Input data contains $N$ images and reports in every batch; a diagnostic report may contain multiple disease descriptions, so $N$ images correspond to $M$ structured labels $l_j, j=1,2,..., M$.
The number of structured labels corresponding to each image is $c_i, i=1,2,...,N, \sum_{i = 1}^N c_i = M $, the $i$-th image corresponds to $c_i$ structured labels $\{l_j | \sum\limits_{k < i} c_k < j \leq \sum\limits_{k \leq i} c_k\}$.
The semantic similarity between the $M$ structured labels is calculated to obtain a matrix of size $M\times M$. Then the maximum values of each column in a sub-matrix of size $c_i\times M$ are compressed to obtain a vector of size $1\times M$, which represents the similarity between the $i$-th medical image and the M structured labels.
During compression, $l_{t^*}$ selects, from a set of $c_i$ structured labels, the one label that has the highest similarity to the $j$-th structured label.
\begin{align}
  t^* & = \underset{\sum\limits_{k < i}c_k < t \leq \sum\limits_{k \leq i}c_k}{Argmax} (cos(E_{sim}(l_t),E_{sim}(l_j))),
\end{align}%

where $E_{sim}$ is the similarity model that is pre-trained by medical knowledge and used to obtain the feature of the text.
Applying max pooling to compress the sub-matrix is well-suited for multi-label image classification by preserving the strongest associations between each image and potential labels.
We do not use average pooling because we want to avoid including weak matching in the calculation and prevent lowering the weights of strong matching.

The label matrix is $s$, where $s_{ij}$ represents the similarity between the $i$-th image and the $j$-th structured label, and the size of $s$ is $N\times M$.
The calculation formula for $s$ is:
\begin{align}
  s_{ij} & = cos(E_{sim}(l_{t^*}),E_{sim}(l_j)),
\end{align}%
where $s_{ij}$ is the maximum cosine similarity between $c_i$ structured labels and the $j$-th structured label as the similarity between the $i$-th image and the $j$-th structured label. The matrix $s$ will be used as the label matrix for vision-language contrastive learning, as shown in Fig.~\ref{fig:2}.

\textbf{Summarized advantages.}
The Semantic Similarity Matrix (SSM) enhances medical image analysis by supporting multi-label classification and providing a more granular similarity measure between images and disease labels.
Its use leads to improved model performance through better visual-language alignment and detailed semantic representation.

\subsection{Multimodal Contrastive Learning}

\paragraph{Training}
The image and text are processed through their respective feature extractors:
\begin{align}
  i=f_{img}(E_{img}(x_{img})),
\end{align}%
\begin{align}
  l=f_{txt}(E_{txt}(x_{txt})),
\end{align}%
where $x_{img}$ and $x_{txt}$ represent the image and text inputs, $E_{img}$ and $E_{txt}$ are the encoders that extract image-specific features and distill medical knowledge from text, respectively.
$f_{img}$ and $f_{txt}$ are projections used to map image and text features into the same embedded space, producing the final features $i$ and $l$.
Then a cosine similarity is computed to predict the similarity between images and texts:
\begin{align}
  Y_{i,j}=\frac{I_i^{\top } \cdot L_j}{||I_i^{\top }|| \cdot ||L_j||} ,
\end{align}%
where $I$ is the set of image feature vectors, $L$ is the set of text feature vectors, and $Y$ is the cosine similarity between images and texts, which is obtained by computing the inner product of $I$ and $L$.

The model adjusts the proximity of embeddings in this shared space based on label similarity; the higher the similarity, the closer the embeddings, as shown in Fig.~\ref{fig:4}.
Embeddings from similar image and text are positioned closer together in the embedded space, embeddings from different image and text are positioned farther.
Due to the high similarity between subclasses, yet not being identical, clustering of subclasses will be carried out within the same class.

\paragraph{Image-Text Matching Loss (ITM)}
ITM loss refines the model's learning process by treating similarity as a continuous variable.
It combines the mean squared error (MSE) with cross-entropy (CE) loss to better capture the degree of association between images and texts:
\begin{align}
  \mathcal{L} (y, s) = \underbrace{\frac{1}{nm}  \sum _{i=1}^{n} \sum _{j=1}^{m}(y_{i,j} - s_{i,j})^2}_{MSE Loss}
  \underbrace{- \frac{1}{n} \sum _{i=1}^{n} \sum _{j=1}^{m} y_{i,j}log(s_{i,j})}_{CE Loss},
\end{align}%
where $y$ is predicted matrix, $s$ is the sementic similarity matrix, $n$ represents the number of images, $m$ represents the number of fine-grained entities.

\paragraph{Inference}
By calculating the similarity between image embedding and text embedding, the model is able to determine whether an image $I$ belongs to a certain category $C$,
\begin{align}
  P(I\in C) = Sim(E_{img}(I), E_{txt}(template(C,Dict(C)))),
\end{align}%
\begin{align}
  C^* & = \underset{C \in V}{Argmax} (P(I\in C)),
\end{align}%
where $Dict$ can map $C$ to its explanation, and $C$ will be enhanced by a template:``\{severity\}+\{location\}+\{category\}, where \{category\} is \{explanation\}''.
$E_{img}$ and $E_{txt}$ represent the encodings of the text and image, $Sim$ calculates the similarity between the text and image embeddings, and the similarity score represents the probability of the image belonging to the category $C$. $V$ is the category library, by comparing the image with all categories in the category library, the most similar $C^*$ to the image can be obtained.

\textbf{Summarized advantages.}
We use a dual-encoder framework to extract features from medical images and texts, then a precise alignment between images and textual descriptors is established, and images can be classified by comparing the computed similarity scores against a comprehensive library of textual explanations.

\section{Experiments}

\begin{table*}[htbp]
  \centering
  \caption{Zero-shot Classification Results}
  \begin{tabular}{l|cc|cc|cc|cc|cc|cc}
    \toprule
    Dataset & \multicolumn{2}{c|}{MIMIC-16} & \multicolumn{2}{c|}{Pneumonia} & \multicolumn{2}{c|}{MIMIC-multi}
            & \multicolumn{2}{c|}{ChestX-ray14} & \multicolumn{2}{c|}{VinBigData} & \multicolumn{2}{c}{MIMIC-FG} \\
    Metric  & ACC$\uparrow$ & F1$\uparrow$ & ACC$\uparrow$ & F1$\uparrow$ & ACC$\uparrow$ & F1$\uparrow$
            & ACC$\uparrow$ & F1$\uparrow$ & ACC$\uparrow$ & F1$\uparrow$ & ACC$\uparrow$ & F1$\uparrow$ \\
    \midrule
    CLIP    & 12.26 & 20.30 & 51.23 & 31.55 & 11.83 & 18.90 & 18.32 & 15.44 & 13.79 & 16.13 & 26.01 & 19.82 \\
    ConVIRT & 63.90 & 26.28 & 76.11 & 58.42 & 54.63 & 26.76 & 71.02 & 16.28 & 53.05 & 23.01 & 54.91 & 31.07 \\
    GloRIA  & 62.54 & 21.47 & 71.29 & 49.01 & 57.40 & 24.60 & 77.00 & 17.32 & 55.20 & 21.87 & 53.92 & 28.95 \\
    MedCLIP & 64.48 & 25.05 & 76.82 & 59.63 & 59.23 & 31.65 & 72.48 & 19.94 & 58.04 & 27.46 & 59.77 & 32.55 \\
    MedKLIP & 69.67 & 32.38 & 80.02 & 63.42 & 64.37 & 36.87 & \textbf{86.19} & 25.25 & 67.87 & 30.78 & 65.44 & 36.83 \\
    \midrule
    Ours    & \textbf{74.71} & \textbf{37.53} & \textbf{81.23} & \textbf{66.08} & \textbf{71.06} & \textbf{40.20}
            & 85.67 & \textbf{28.78} & \textbf{70.03} & \textbf{38.69} & \textbf{71.85} & \textbf{41.27} \\
    \bottomrule
  \end{tabular}
  \label{tab:1}
\end{table*}

\subsection{Experimental Setup}
\paragraph{Datasets}
\textbf{RSNA-Pneumonia}~\cite{shih2019augmenting}, contains around 30k chest X-ray images including normal and pneumonia cases.
We use 1,024 samples to test the model's capability for single-label classification.

\textbf{NIH ChestX-ray14}~\cite{wang2017chestx}, contains around 112k chest X-ray images of 18 chest diseases.
We use 1,024 samples containing 10 common diseases to test the model's capability for multi-label classification.

\textbf{VinBigData}~\cite{nguyen2022vindr}, contains around 18k chest X-ray images of 14 chest diseases.
We use 1,024 samples containing 8 common diseases to test the model's capability for multi-label classification.

\textbf{COVID}~\cite{rahman2021exploring}, contains around 18k chest X-ray images including non-COVID and COVID cases.
We use 1,024 samples to test the model's performance on unseen types.

\textbf{MIMIC-CXR}~\cite{goldberger2000physiobank,johnson2019mimic}, contains around 370k chest X-ray images paired with associated radiology reports.
We obtain four datasets from the MIMIC-CXR:

\textbf{MIMIC-train}, contains 50 diseases with a positive ratio of over 0.0001, labeled as severity + lesion location + category, with 50,000 samples for model pre-training;

\textbf{MIMIC-16}, contains 16 diseases with a positive ratio of over 0.01 to test the model's capability for single-label classification;

\textbf{MIMIC-multi}, contains 16 diseases with a positive ratio of over 0.01 to test the model's capability for multi-label classification;

\textbf{MIMIC-FG}, contains the 5 most common diseases and 3 disease severity, with each sample potentially containing multiple diseases and severity + category as labels, to test the model's fine-grained classification performance.

\paragraph{Baselines}
\textbf{ResNet}\cite{resnet} is a deep convolutional neural network architecture that utilizes residual connections to enable the training of very deep networks by mitigating the vanishing gradient problem.

\textbf{ConvNeXt}\cite{convnext} is a modernized convolutional neural network architecture that incorporates design principles from transformer models to enhance performance and efficiency in computer vision tasks.

\textbf{CLIP}\cite{radford2021clip} performs image-text contrastive learning through natural language supervision signals.

\textbf{ConVIRT}\cite{zhang2022convirt} performs contrastive learning on medical image-report paired data.

\textbf{GloRIA}\cite{huang2021gloria} contrasts global image representations with reports and local image patches with report words.

\textbf{MedCLIP}\cite{wang2022medclip} decouples image-text pairs using NER for multi-modal contrastive learning, addressing issues of false negatives and lack of training data.

\textbf{MedKLIP}\cite{wu2023medklip} extracts disease information from diagnostic reports and augments it with medical domain knowledge to improve visual language processing in radiology.

\paragraph{Implementation Details}
We employed bioClinicalBERT \cite{biobert_2019} for semantic similarity, ResNet50 \cite{resnet} as the image encoder, and Transformer \cite{transformer} as the text encoder; both encoders are initialized with CLIP pre-trained weights.
The learning process utilized a dual loss function strategy: cross-entropy loss for contrastive learning and mean squared error loss for precise matching in fine-grained classification tasks.
Stochastic gradient descent (SGD) was chosen as the optimizer, starting with a learning rate of 0.006 and using Cosine Annealing Warm Restarts for learning rate adjustment.
During data preprocessing, the images were cropped to a single-channel size of 448x448 pixels.
Standardization was applied post-cropping without further augmentations.
The pretraining phase was executed on a single RTX-3090 GPU, spanning 100 epochs and taking roughly 10 hours to complete.

\subsection{Downstream Tasks}

\subsubsection{Zero-shot Classification}
\begin{table*}[t]
  \centering
  \caption{Fine-Tuning Classification Results With Different Portions of Training Data}
  \label{tab:2}
  \begin{tabular}{l|cc|cc|cc|cc|cc|cc}
    \toprule
    Dataset & \multicolumn{6}{c|}{Pneumonia} &  \multicolumn{6}{c}{VinBigData} \\
    \midrule
    Portion & \multicolumn{2}{c|}{1\%} & \multicolumn{2}{c|}{10\%} & \multicolumn{2}{c|}{100\%} & \multicolumn{2}{c|}{1\%} & \multicolumn{2}{c|}{10\%} & \multicolumn{2}{c}{100\%} \\
    Metric  & ACC$\uparrow$ & F1$\uparrow$ & ACC$\uparrow$ & F1$\uparrow$ & ACC$\uparrow$ & F1$\uparrow$ & ACC$\uparrow$ & F1$\uparrow$ & ACC$\uparrow$ & F1$\uparrow$ & ACC$\uparrow$ & F1$\uparrow$ \\
    \midrule
    Resnet50 (Random Init) & 63.48 & 51.43 & 74.54 & 58.65 & 79.39 & 64.71 & 60.35 & 29.39 & 65.81 & 33.92 & 67.20 & 36.71\\
    ConvNeXt (Random Init) & 69.53 & 49.65 & 75.00 & 62.57 & 78.13 & 63.42 & 59.48 & 28.15 & 66.87 & 33.83 & 71.61 & 36.42\\
    \midrule
    Resnet50 (ImageNet Init) & 78.52 & 62.17 & 77.05 & 66.37 & 80.47 & 68.04 & 60.28 & 34.89 & 68.34 & 37.05 & 68.70 & 39.10\\
    ConvNeXt (ImageNet Init) & 77.34 & 64.06 & 79.12 & 64.07 & 78.32 & 68.82 & 62.39 & 33.72 & 67.23 & 36.40 & 70.29 & 37.93\\
    \midrule
    MedCLIP (MIMIC-CXR Pre-trained)& 80.11 & 64.62 & 80.27 & 66.41 & 82.31 & 70.36 & 67.39 & 34.14 & 70.38 & 38.79 & 73.80 & 40.96\\
    MedKLIP (MIMIC-CXR Pre-trained)& 80.84 & 67.55 & 81.29 & 69.32 & 81.95 & 71.20 & 69.01 & 36.25 & 71.20 & 39.65 & 74.54 & 41.87\\
    \midrule
    Ours    & \textbf{81.59} & \textbf{68.73} & \textbf{83.14} & \textbf{71.69} & \textbf{84.96} & \textbf{72.28} & \textbf{73.95} & \textbf{39.71} & \textbf{75.45} & \textbf{41.81} & \textbf{75.52} & \textbf{44.01}\\
    \bottomrule
  \end{tabular}
\end{table*}

\paragraph{Experimental settings}

The classification results are based on the candidate texts with a cosine similarity exceeding the threshold to the input image. For our approach, the candidate texts consist of image-specific explanations related to the diseases. In contrast, the candidate texts used by previous methods are categories combined with a prompt template, following the format “This is an X-ray image of \{category\}.” For example, the MIMIC-16 dataset includes 16 types of diseases, and the candidate texts corresponding to these 16 diseases are used to calculate the cosine similarity with the image. The disease categories with cosine similarity exceeding a certain threshold are selected as the classification results for the image. We use accuracy and F1 score to validate the effectiveness of the method. Accuracy measures the fraction of correctly predicted instances out of the total instances and F1 score is the harmonic mean of precision (the accuracy of positive predictions) and recall (the ability to find all positive instances). In medical datasets, the positive and negative class samples are highly imbalanced; thus, in this paper, we use the threshold that maximizes the F1 score, so the accuracy metric is the accuracy when the F1 score is the highest. For datasets that contain multiple diseases, we separately calculate F1 Score and accuracy for each disease and then calculate the average.

\paragraph{Experimental results} MedFILIP demonstrates highly competitive results in zero-shot scenarios across six datasets, outperforming current state-of-the-art (SOTA) image-text pretraining models, Table~\ref{tab:1} shows these findings. Our proposed method demonstrates exceptional robustness and broad applicability, as evidenced by its impressive zero-shot performance across multiple datasets while consistently delivering high-quality results. On MIMIC-16 and RSNA-Pneumonia, we compared single-label classification performance with improvements of 5.04\% and 1.21\%, respectively. On MIMIC-multi, ChestX-ray14, and VinBigData datasets, we compared multi-label classification performance, with improvements by a maximum of 6.69\%. 
MedFILIP's performance metrics on fine-grained classification, which involves distinguishing between subclasses, are superior to other models that only consider broader disease categories.
On the MIMIC-FG dataset that is specifically designed for evaluating fine-grained image classification, we compared fine-grained classification performance with improvements of 6.41\%, as shown in the MIMIC-FG column in Table~\ref{tab:1}.

\begin{table}[t]
  \centering
  \caption{Results of Predictions on Unseen Category}
  \begin{tabular}{l|cc|cc}
    \toprule
    Method  & \multicolumn{2}{c|}{Category} & \multicolumn{2}{c}{Explanation} \\
    Metric  & ACC$\uparrow$ & F1$\uparrow$ & ACC$\uparrow$ & F1$\uparrow$ \\
    \midrule
    MedCLIP & 60.46 & 47.31 & 61.03 & 57.44 \\
    MedKLIP & \textbf{63.38} & 63.26 & 68.57 & 71.43 \\
    \midrule
    Ours    & 62.01 & \textbf{64.77} & \textbf{71.26} & \textbf{72.31} \\
    \bottomrule
  \end{tabular}
  \label{tab:3}
\end{table}

\subsubsection{Fine-tuned Classification}
\paragraph{Experimental settings}
MedFILIP, along with multimodal pre-training methods, was evaluated through linear probing, where the weights of the pre-trained image encoder remained frozen, and a trainable linear classifier was added to generate classification results. For comparison, traditional methods such as ResNet and ConvNeXt were employed with random-initialized parameters or ImageNet-initialized parameters and underwent full parameter training. During fine-tuning, we used 1\%, 10\%, and 100\% of the data to assess the data efficiency of visual representation learning.

\paragraph{Experimental results}
MedFILIP demonstrates exceptional performance in conducting fine-tuned validation across two distinct datasets: RSNA-Pneumonia and VinBigData. As shown in Table~\ref{tab:2}, our method surpasses random-initialized or ImageNet-initialized non-pretraining approaches while utilizing only 10\% of the data compared to the full dataset used by others, which indicates that our pre-training methods enhance image feature representation capabilities, which can reduce the reliance on labeled data for downstream tasks. Meanwhile, our method outperforms methods that are also pre-trained on the MIMIC-CXR dataset, highlighting the advantage of pairing fine-grained entities with images and injecting image-specific and medical domain knowledge into the model.

\subsubsection{Unseen Type Classification}
\paragraph{Experimental settings}
To evaluate our model's applicability across a broad spectrum, we assessed it on an unseen disease under zero-shot. In Table~\ref{tab:3}, the term ``Category" signifies the use of the known type ``Pneumonia" to denote COVID-19 during the prediction phase. Simultaneously, ``Explanation" stands for the utilization of image-specific explanations to represent COVID-19. The explanation of COVID-19 is \textit{Bilateral, diffuse, or localized areas of shadowing or opaque lesions in the lung periphery and lower lobes, appearing gray and hazy with blurred borders.} 

\paragraph{Experimental results}
Table~\ref{tab:3} illuminates MedFILIP's potent capability in predicting unseen disease categories. 
Traditional machine learning models reliant solely on historical data fall short in recognizing COVID-19, as it was not present in sets of data predating 2019. Nonetheless, our model, MedFILIP, can leverage the semantic information embedded in textual descriptions conveying the attributes of COVID-19 and make appropriate predictions. Despite the absence of training data labeled as COVID-19, the model can recognize related features such as ``bilateral" and ``shadowing." As a result, when presented with a textual description of symptoms synonymous with COVID-19, the model can evaluate whether a provided image aligns with this description, thereby determining the probability of the image suggesting a COVID-19 case. 

\begin{table}[t]
  \centering
  \caption{Dice Score of Segmentation With Different Portions of Fine-Tuning Data}
  \label{tab:4}
  \begin{tabular}{l|p{0.95cm}|p{0.95cm}|p{0.95cm}}
    \toprule
    Dataset & \multicolumn{3}{c}{Pneumothorax Segmentation} \\
    \midrule
    Portion & \enspace 1\% & \enspace 10\% & \enspace 100\% \\
    \midrule
    Random & \enspace 23.27 & \enspace 49.30 & \enspace 66.04 \\
    MedCLIP & \enspace 60.83 & \enspace 69.58 & \enspace 75.27 \\
    MedKLIP & \enspace 66.59 & \enspace 72.10 & \enspace 79.37\\
    \midrule
    Ours    & \enspace \textbf{68.51} & \enspace \textbf{77.42} & \enspace \textbf{80.65}\\
    \bottomrule
  \end{tabular}
\end{table}

\begin{table}[t]
  \centering
  \caption{Results of Image-to-Text Retrieval and Text-to-Image Retrieval}
  \label{tab:5}
  \begin{tabular}{l|c|c|c|c|c|c}
    \toprule
    Task & \multicolumn{3}{c|}{Image to Text} & \multicolumn{3}{c}{Text to Image} \\
    \midrule
    Metric & P@1 & P@3 & P@10 & P@1 & P@3 & P@10 \\
    \midrule
    ConVIRT & 9.92 & 18.59 & 47.16 & 12.33 & 22.10 & 44.07\\
    GloRIA & 11.45 & 19.26 & 49.93 & 14.54 & 23.65 & 49.24 \\
    MedCLIP & 10.20 & 19.74 & 50.62 & 13.25 & 23.49 & 50.05 \\
    MedKLIP & 11.18 & 20.35 & 51.27 & 15.50 & 24.28 & 53.36 \\
    \midrule
    Ours    & \textbf{14.01} & \textbf{24.39} & \textbf{53.75} & \textbf{16.11} & \textbf{27.83} & \textbf{56.63} \\
    \bottomrule
  \end{tabular}
\end{table}

\begin{table*}[ht]
  \centering
  \caption{Ablation studies}
  \begin{tabular}{cccc|cc|cc|cc|cc|cc|cc}
    \toprule
    \multirow{2}{*}{SSM} & \multirow{2}{*}{IKI} & \multirow{2}{*}{FG} & \multirow{2}{*}{Entity} & \multicolumn{2}{c|}{MIMIC-16} & \multicolumn{2}{c|}{Pneumonia} & \multicolumn{2}{c|}{MIMIC-multi} & \multicolumn{2}{c|}{ChestX-ray14} & \multicolumn{2}{c|}{VinBigData} & \multicolumn{2}{c}{MIMIC-FG} \\
    ~ & ~ & ~ & ~ & ACC$\uparrow$ & F1$\uparrow$ & ACC$\uparrow$ & F1$\uparrow$ & ACC$\uparrow$ & F1$\uparrow$ & ACC$\uparrow$ & F1$\uparrow$ & ACC$\uparrow$ & F1$\uparrow$ & ACC$\uparrow$ & F1$\uparrow$ \\
    \midrule
    
    & & & & 63.9 & 26.3 & 76.1 & 58.4 & 54.6 & 26.8 & 71.0 & 16.3 & 53.1 & 23.0 & 54.9 & 31.1 \\
    & & & $\checkmark$ & 67.2 & 29.4 & 75.6 & 57.6 & 61.8 & 33.6 & 76.9 & 21.3 & 61.9 & 27.2 & 58.0 & 31.6 \\
    & & $\checkmark$ & $\checkmark$ & 70.6 & 32.6 & 77.3 & 60.7 & 65.0 & 35.7 & 79.2 & 24.5 & 66.0 & 31.5 & 66.3 & 36.1 \\
    $\checkmark$ & & $\checkmark$ & $\checkmark$ & 73.3 & 35.4 & 80.0 & 62.2 & 69.8 & 39.8 & 84.2 & 28.3 & 68.4 & 36.2 & 71.7 & 40.7 \\
    $\checkmark$ & $\checkmark$ & $\checkmark$ & $\checkmark$ & \textbf{74.7} & \textbf{37.5} & \textbf{81.2} & \textbf{66.1} & \textbf{71.1} & \textbf{40.2} & \textbf{85.7} & \textbf{28.8} & \textbf{70.0} & \textbf{38.7} & \textbf{71.9} & \textbf{41.3} \\
    \bottomrule
  \end{tabular}
  \label{tab:6}
\end{table*}

\subsubsection{Segmentation}

\paragraph{Experimental settings}
SIIM ACR Pneumothorax~\cite{pneumothorax} dataset was used to evaluate the model's performance when transferred to segmentation tasks. This dataset contains around 12k chest X-ray images with masks of pneumothorax. We fine-tuned the UNet for the segmentation. Our approach, along with previous pre-training methods, employed pre-trained vision encoders to initialize the feature extraction layers of the UNet. A randomly initialized UNet served as a comparative baseline. We fine-tuned the models using 1\%, 10\%, and 100\% of the data to evaluate the model's data utilization efficiency. We used Dice Score to validate the effectiveness of the method. Dice Score measures the overlap between the predicted segmentation image and the ground truth image.

\paragraph{Experimental results}

MedFILIP aids in learning semantic and visual features, contributing to improved performance in segmentation tasks, as shown in Table~\ref{tab:4}. Fine-tuning the UNet using our method's pre-trained weights significantly reduces training time and surpasses prior methods with only 10\% of the data compared to the full dataset used by others. The efficiency indicates the model's acquisition of substantial foundational knowledge, necessitating fewer adjustments for the specific task.

\subsubsection{Retrieval}

\paragraph{Experimental settings}

We randomly selected 1,000 samples from the MIMIC-CXR dataset, each associated with a specific disease. These samples were transformed into image-text pairs. For our approach, the texts were image-specific explanations related to the diseases. In contrast, texts used in previous methods were categories combined with a prompt template, following the format “This is an X-ray image of \{category\}.” Pre-trained vision-language models can be applied to retrieval tasks without additional fine-tuning. For instance, when retrieving the text most similar to a given image from 1,000 texts, the image and all texts were processed through their respective encoders to generate feature vectors. The cosine similarity between the image and text feature vectors was then calculated. The text with the highest cosine similarity to the image was considered the most relevant match. We used Precision@K to validate the effectiveness of the method. Precision@K measures how many of the top K items retrieved by the model are relevant or correct.

\paragraph{Experimental results}

MedFILIP achieves more accurate content matching by simultaneously comprehending both images and texts, as shown in Table~\ref{tab:5}. Our model demonstrates superior performance in both image-to-text and text-to-image retrieval with improvements by a maximum of 4.04\%, which indicates its versatility and strength in processing multimodal information, further broadening its application scope in clinical contexts.

\subsection{Ablation Study}
\paragraph{Experimental settings}

As shown in Table~\ref{tab:6}, if Entity is enabled, the image-text alignment will be performed between images and disease entities; otherwise, the alignment will be performed between images and reports.
If FG is enabled, the entities will contain detailed disease triplets; otherwise, the entities will only contain disease categories.
If IKI is enabled, medical visual domain knowledge will be used to enhance the texts; otherwise, the texts will only contain disease triplets.
If SSM is enabled, the label matrix will be calculated by the similarity between texts based on extracted entities and image-specific explanations; otherwise, the label matrix would only contain 0s or 1s.

\paragraph{Experimental results}

An extensive ablation study was carried out to confirm the efficacy of different components of our MedFILIP method, as illustrated in Table~\ref{tab:6}. In comparing image-entity alignment with image-report alignment (comparing row 1 with row 2), the findings demonstrate that aligning images directly with disease entities yields greater precision, enhancing the accuracy of image-text pairing. The superiority of fine-grained entities (comparing row 2 with row 3) is evident as they provide more detailed disease information, significantly improving alignment efficacy. Moreover, fine-grained entities not only retain the majority of essential information but also facilitate more precise image pairing(comparing row 1 with row 3). Implementing an SSM (comparing row 3 with row 4) enhances the label matrix’s expressiveness, further boosting model performance. Similarly, the introduction of medical visual domain knowledge (IKI) yields notable improvements (comparing row 4 with row 5), enriching the textual descriptions and thereby augmenting the model’s interpretative capabilities, which in turn improves alignment.

\begin{figure*}[t]
  \centering
  \subfigure{
    \includegraphics[width=\columnwidth]{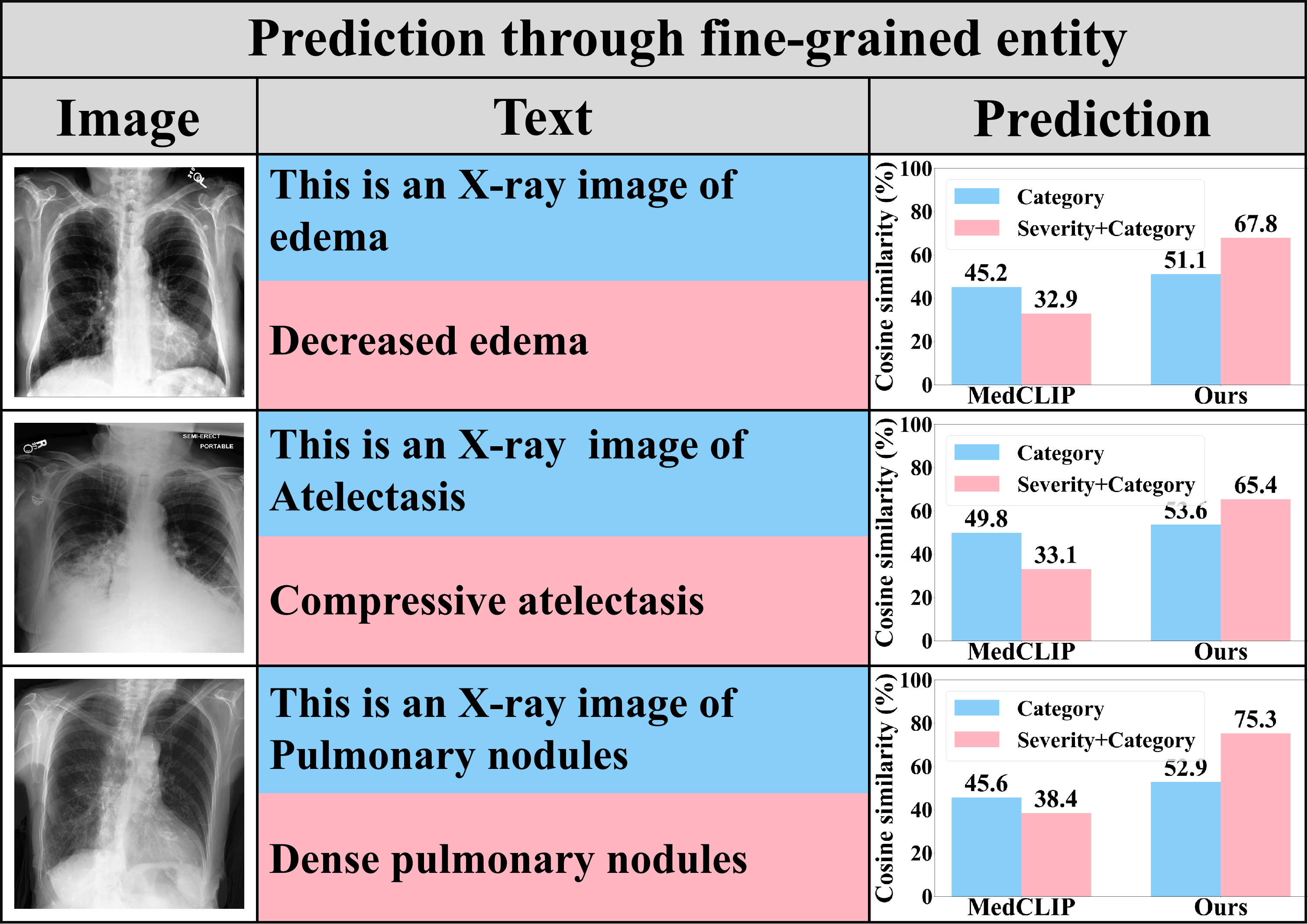}}
  \subfigure{
    \includegraphics[width=\columnwidth]{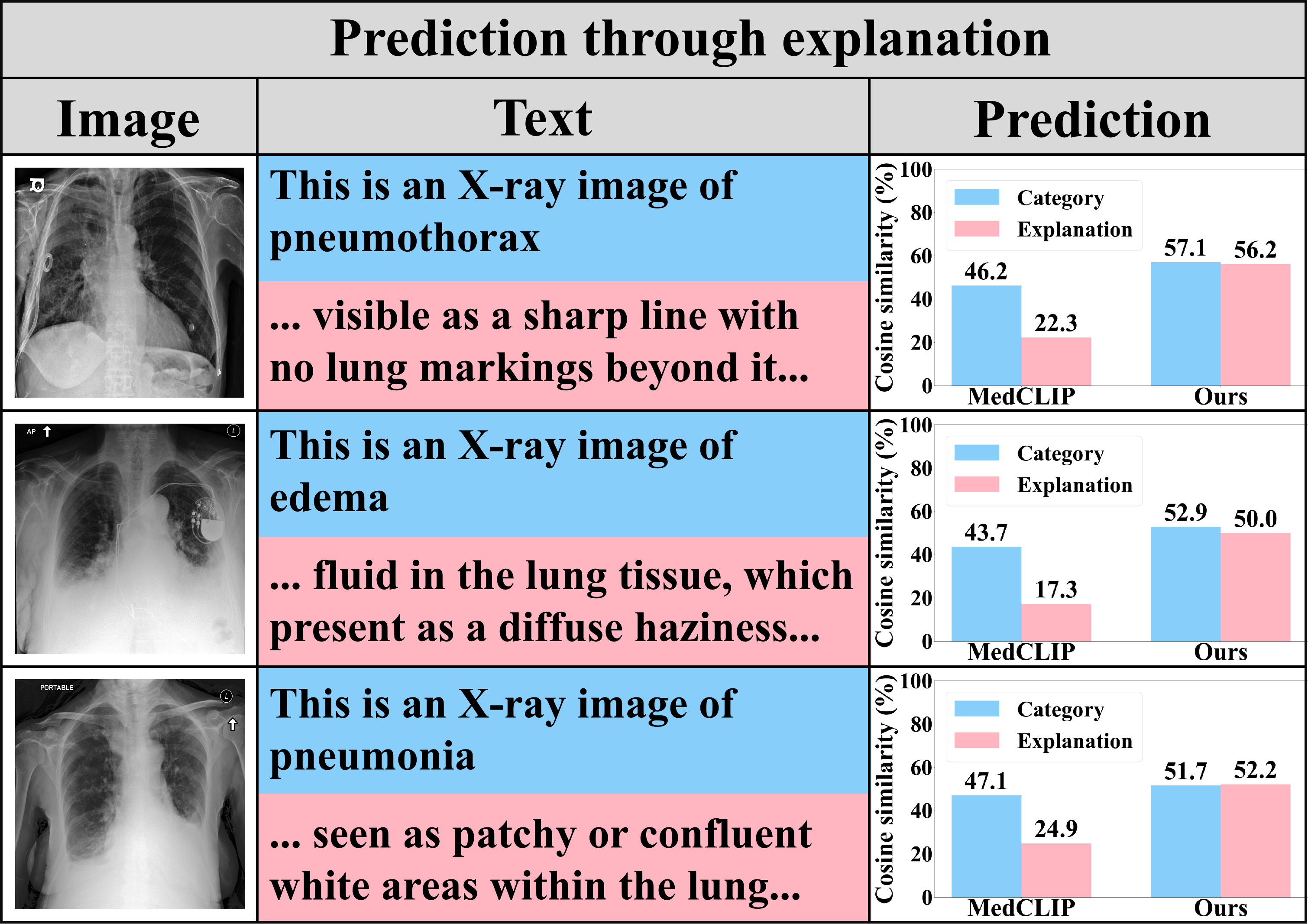}}
  \DeclareGraphicsExtensions.
  \caption{
    A comparison of the predicted cosine similarity scores between images and texts demonstrates that
    MedFILIP is better at classifying subclasses and associating images with fine-grained explanations. There are two parts of cosine similarity scores between the images and the corresponding texts, the left part calculates the similarities between the images and the texts of "template + category" as well as the texts of fine-grained categories, and the right part calculates the similarities between the images and the texts of "template + category" as well as the texts of fine-grained explanations of the image. In the bar chart of similarity scores, the bars correspond to the text segments highlighted in the same colors, the blue bars are cosine similarity between the images and the texts of "template + category", and the pink bars are cosine similarity between the images and the texts of fine-grained categories or explanations.
  }
  \label{fig:6}
\end{figure*}

\subsection{Case Study}
Fig.~\ref{fig:6} evidences the promising performance of MedFILIP for matching images with both fine-grained entities and image-specific explanations. The pre-training models employ contrastive learning between images and text to acquire rich multimodal representations in the training phase, which enables the model to discern the similarity between different images and texts effectively. In the inference phase, the classification output of the pre-training models is determined by selecting the candidate text with the highest cosine similarity to the input image, thereby determining whether the image belongs to a certain type of disease. To assess the pre-training models' representation capability, we compute similarity scores between paired images and texts. A high similarity score signifies that the model can effectively extract key features and meanings from both the images and texts, and establish connections across the different modalities. In this study, the images are chest X-ray images, and the texts are of three types: (1) a prompt template combined with disease category, using the format ``This is an X-ray image of \{category\}"; (2) fine-grained disease categories, incorporating disease severity and disease category; (3) explanations of the visual attributes and medical knowledge related to the disease. Previous approaches achieve higher image-text similarity using the prompt template + category method, whereas our approach attains higher similarity with the latter two text types. Our method demonstrates a superior predictive correlation between images and their fine-grained labels compared to the correlation between images and their general category labels. This result diverges from those obtained by earlier methods, highlighting advancements in our approach for inferring fine-grained disease categories. When replacing basic category labels with image-specific explanations to associate with the images, our method achieves higher similarity scores, indicating its enhanced effectiveness in establishing relationships between images and their specific, image-related explanations.

\section{Discussion}
The proposed approach mainly solves the following two types of challenges, which are crucial for the advancement of medical vision-language pretraining:

\textbf{Concurrent diseases and repeated occurrences.} A diagnostic report can encompass multiple disease descriptions and a specific disease description may recur across various reports. Direct usage of the comprehensive report may introduce the risk of entangling separate disease descriptions or could result in images of identical diseases being misclassified into different categories. In plain terms, these problems arise due to the complexity embedded in the text.

\textbf{Multiple subclasses and unseen class.} A disease category may consist of various subclasses, each presenting slightly dissimilar symptoms. For instance, different stages of cancer may exhibit unique visual characteristics. Sole reliance on training labels, that focus merely on broad disease categories, may inadvertently overlook the intricate distinctions distinct to each subclass. Moreover, an unforeseen class, unrepresented in the training phase, may emerge during prediction, such as COVID-19, absent from data sets pre-2019 and, thus, inaccurately predicted by prior models due to a lack of COVID-specific information. These issues arise primarily because the supervising text data lacks an adequate depth of information.

Excessive complexity or too many irrelevant details in the text supervision information can make the model learn inaccurate patterns or associations. Consequently, the accurate learning of image-text representation becomes daunting, leading to potential misclassifications. On the flip side, if the text supervision information lacks richness, the model might fall short in handling complex or novel scenarios, thereby posing challenges for fine-grained learning and generalization. 

Our proposed MedFILIP successfully reduces text complexity while preserving the richness of textual supervision information. It effectively reframes the information extraction task as a text generation task, well-aligned with LLM, skillfully uncoupling disease entities from their respective reports. It also utilizes critical disease details, such as severity and location, to enable precise and fine-grained image-text alignment. Moreover, MedFILIP facilitates knowledge transfer from familiar classes to novel classes sharing similar visual attributes to enable zero-shot on unseen classes. MedFILIP strategically progresses the supervision from a high level of information richness yet high complexity (full diagnostic report) to moderate information richness and reduced complexity (fine-grained entities), finally landing on high information richness but low complexity (fine-grained entities accompanied by image-specific explanations). A high degree of information richness provides the model with a more granular visual-semantic representation through language supervision, whereas a lower complexity eases the training process. Therefore, MedFILIP, by thoughtfully maintaining a balance between text complexity and information richness, outperforms other approaches in terms of medical vision-language pretraining.

MedFILIP, despite holding substantial promise, also presents potential limitations, which will form the focal point of our future research. For instance, although harnessing LLM for automatic information extraction is more efficient and adaptive than manual processes, given the "black box" nature of LLM, fine-tuning these models poses a significant challenge. We aspire to address these issues by crafting more effective prompts and utilizing more robust LLM. Additionally, current explanations are specific to categories and lack details for subclasses. For instance, the generic explanation given for pneumonia is applied across all its subclasses such as severe pneumonia or mild pneumonia. In the future, we expect to create an image-specific explanation database for disease subclasses and build an inference pipeline from image-specific attributes to disease subclasses.

\section{Conclusion}
In this paper, we propose a novel method called MedFILIP that leverages fine-grained annotations to improve medical vision-language contrastive learning. Our proposed MedFILIP effectively circumvents the issues arising from the complexity of diagnostic reports and tackles the challenges in predicting subclasses and unseen classes. The proposed GPT-IE effectively extracts fine-grained entities from diagnosis reports by a large language model, decoupling disease information, including severity, location, and categories from reports, thereby allowing the distinction of disease subclasses. Our IKI maps diseases to fine-grained image-specific explanations, enabling the model to learn relationships between diseases and their attributes, thus allowing predictions on unseen classes based on the knowledge learned from seen classes. Our SSM finally attains labels representing the correlation between images and text based on the aforementioned fine-grained annotations, thereby proceeding with fine-grained image-text comparative learning. A full set of experiments were carried out which demonstrated the proposed MedFILIP framework achieved its goal of enhancing multimodal learning via fine-grained annotations.

\bibliographystyle{IEEEtran}
\bibliography{medfilip}

% Generated by IEEEtran.bst, version: 1.14 (2015/08/26)
\begin{thebibliography}{10}
\providecommand{\url}[1]{#1}
\csname url@samestyle\endcsname
\providecommand{\newblock}{\relax}
\providecommand{\bibinfo}[2]{#2}
\providecommand{\BIBentrySTDinterwordspacing}{\spaceskip=0pt\relax}
\providecommand{\BIBentryALTinterwordstretchfactor}{4}
\providecommand{\BIBentryALTinterwordspacing}{\spaceskip=\fontdimen2\font plus
\BIBentryALTinterwordstretchfactor\fontdimen3\font minus \fontdimen4\font\relax}
\providecommand{\BIBforeignlanguage}[2]{{%
\expandafter\ifx\csname l@#1\endcsname\relax
\typeout{** WARNING: IEEEtran.bst: No hyphenation pattern has been}%
\typeout{** loaded for the language `#1'. Using the pattern for}%
\typeout{** the default language instead.}%
\else
\language=\csname l@#1\endcsname
\fi
#2}}
\providecommand{\BIBdecl}{\relax}
\BIBdecl

\bibitem{boecking2022making}
B.~Boecking, N.~Usuyama, S.~Bannur, D.~C. Castro, A.~Schwaighofer, S.~Hyland, M.~Wetscherek, T.~Naumann, A.~Nori, J.~Alvarez-Valle \emph{et~al.}, ``Making the most of text semantics to improve biomedical vision--language processing,'' in \emph{Lect. Notes Comput. Sci.}\hskip 1em plus 0.5em minus 0.4em\relax Springer, 2022, pp. 1--21.

\bibitem{li2024mlip}
Z.~Li, L.~T. Yang, B.~Ren, X.~Nie, Z.~Gao, C.~Tan, and S.~Z. Li, ``Mlip: Enhancing medical visual representation with divergence encoder and knowledge-guided contrastive learning,'' in \emph{Proc. IEEE Conf. Comput. Vis. Pattern Recognit.}, 2024, pp. 11\,704--11\,714.

\bibitem{liu2023improving}
B.~Liu, D.~Lu, D.~Wei, X.~Wu, Y.~Wang, Y.~Zhang, and Y.~Zheng, ``Improving medical vision-language contrastive pretraining with semantics-aware triage,'' \emph{IEEE Trans. Med. Imaging}, vol.~42, no.~12, pp. 3579--3589, 2023.

\bibitem{zhang2022convirt}
Y.~Zhang, H.~Jiang, Y.~Miura, C.~D. Manning, and C.~P. Langlotz, ``Contrastive learning of medical visual representations from paired images and text,'' in \emph{Proc. Mach. Learn. Res.}\hskip 1em plus 0.5em minus 0.4em\relax PMLR, 2022, pp. 2--25.

\bibitem{wu2023medklip}
C.~Wu, X.~Zhang, Y.~Zhang, Y.~Wang, and W.~Xie, ``Medklip: Medical knowledge enhanced language-image pre-training for x-ray diagnosis,'' in \emph{Proc. IEEE Int. Conf. Comput. Vis.}, 2023, pp. 21\,372--21\,383.

\bibitem{wang2022medclip}
Z.~Wang, Z.~Wu, D.~Agarwal, and J.~Sun, ``Medclip: Contrastive learning from unpaired medical images and text,'' in \emph{Proc. Conf. Empir. Methods Nat. Lang. Process., EMNLP}, 2022, pp. 3876--3887.

\bibitem{irvin2019chexpert}
J.~Irvin, P.~Rajpurkar, M.~Ko, Y.~Yu, S.~Ciurea-Ilcus, C.~Chute, H.~Marklund, B.~Haghgoo, R.~Ball, K.~Shpanskaya \emph{et~al.}, ``Chexpert: A large chest radiograph dataset with uncertainty labels and expert comparison,'' in \emph{AAAI Conf. Artif. Intell., AAAI, Innov. Appl. Artif. .igence Conf., IAAI AAAI Symp. Educ. Adv. Artif. Intell., EAAI}, 2019, pp. 590--597.

\bibitem{johnson2019mimic}
A.~E. Johnson, T.~J. Pollard, S.~J. Berkowitz, N.~R. Greenbaum, M.~P. Lungren, C.-y. Deng, R.~G. Mark, and S.~Horng, ``Mimic-cxr, a de-identified publicly available database of chest radiographs with free-text reports,'' \emph{Sci. Data}, vol.~6, no.~1, p. 317, 2019.

\bibitem{radford2021clip}
A.~Radford, J.~W. Kim, C.~Hallacy, A.~Ramesh, G.~Goh, S.~Agarwal, G.~Sastry, A.~Askell, P.~Mishkin, J.~Clark \emph{et~al.}, ``Learning transferable visual models from natural language supervision,'' in \emph{Proc. Mach. Learn. Res.}\hskip 1em plus 0.5em minus 0.4em\relax PMLR, 2021, pp. 8748--8763.

\bibitem{oakden2020hidden}
L.~Oakden-Rayner, J.~Dunnmon, G.~Carneiro, and C.~R{\'e}, ``Hidden stratification causes clinically meaningful failures in machine learning for medical imaging,'' in \emph{Proc. ACM Conf. Health, Inference, Learn.}, 2020, pp. 151--159.

\bibitem{dai2023efficient}
L.~Dai, W.~Lei, and X.~Zhang, ``Efficient subclass segmentation in medical images,'' in \emph{Lect. Notes Comput. Sci.}\hskip 1em plus 0.5em minus 0.4em\relax Springer, 2023, pp. 266--275.

\bibitem{Flamingo}
J.-B. Alayrac, J.~Donahue, P.~Luc, A.~Miech, I.~Barr, Y.~Hasson, K.~Lenc, A.~Mensch, K.~Millican, M.~Reynolds \emph{et~al.}, ``Flamingo: a visual language model for few-shot learning,'' \emph{Adv. neural inf. proces. syst.}, vol.~35, pp. 23\,716--23\,736, 2022.

\bibitem{LLaVA}
H.~Liu, C.~Li, Q.~Wu, and Y.~J. Lee, ``Visual instruction tuning,'' \emph{Adv. neural inf. proces. syst.}, vol.~36, 2024.

\bibitem{LLaVA-Med}
C.~Li, C.~Wong, S.~Zhang, N.~Usuyama, H.~Liu, J.~Yang, T.~Naumann, H.~Poon, and J.~Gao, ``Llava-med: Training a large language-and-vision assistant for biomedicine in one day,'' \emph{Adv. neural inf. proces. syst.}, vol.~36, 2024.

\bibitem{LLM_CXR}
S.~Lee, W.~J. Kim, J.~Chang, and J.~C. Ye, ``Llm-cxr: Instruction-finetuned llm for cxr image understanding and generation,'' in \emph{Proc. Int. Conf. Learn. Represent.}, 2024, pp. 1--13.

\bibitem{chatcad}
Z.~Zhao, S.~Wang, J.~Gu, Y.~Zhu, L.~Mei, Z.~Zhuang, Z.~Cui, Q.~Wang, and D.~Shen, ``Chatcad+: Towards a universal and reliable interactive cad using llms,'' \emph{IEEE Trans. Med. Imaging}, 2024.

\bibitem{huang2021gloria}
S.-C. Huang, L.~Shen, M.~P. Lungren, and S.~Yeung, ``Gloria: A multimodal global-local representation learning framework for label-efficient medical image recognition,'' in \emph{Proc. IEEE Int. Conf. Comput. Vis.}, 2021, pp. 3942--3951.

\bibitem{FG1}
Y.~Zeng, X.~Zhang, and H.~Li, ``Multi-grained vision language pre-training: Aligning texts with visual concepts,'' in \emph{Proc. Int. Conf. Mach. Learn.}\hskip 1em plus 0.5em minus 0.4em\relax PMLR, 2022, pp. 25\,994--26\,009.

\bibitem{FG3}
M.~Xu, Z.~Zhang, H.~Hu, J.~Wang, L.~Wang, F.~Wei, X.~Bai, and Z.~Liu, ``End-to-end semi-supervised object detection with soft teacher,'' in \emph{Proc. IEEE Int. Conf. Comput. Vis.}, 2021, pp. 3060--3069.

\bibitem{FG4}
S.~Branson, G.~Van~Horn, S.~Belongie, and P.~Perona, ``Bird species categorization using pose normalized deep convolutional nets,'' \emph{arXiv}, 2014.

\bibitem{FG5}
X.~Liu, J.~Wang, S.~Wen, E.~Ding, and Y.~Lin, ``Localizing by describing: Attribute-guided attention localization for fine-grained recognition,'' in \emph{AAAI Conf. Artif. Intell., AAAI}, vol.~31, 2017, pp. 4190--4196.

\bibitem{FG0}
J.~Zuo, J.~Hong, F.~Zhang, C.~Yu, H.~Zhou, C.~Gao, N.~Sang, and J.~Wang, ``Plip: Language-image pre-training for person representation learning,'' in \emph{Proc. Adv. Neural Inf. Process. Syst.}, 2024, pp. 1--15.

\bibitem{Med_know}
X.~Xie, J.~Niu, X.~Liu, Z.~Chen, S.~Tang, and S.~Yu, ``\BIBforeignlanguage{en-US}{A survey on incorporating domain knowledge into deep learning for medical image analysis},'' \emph{\BIBforeignlanguage{en-US}{Med. Image. Anal.}}, p. 101985, Apr 2021.

\bibitem{Med_know2}
I.~Gonzalez-Diaz, ``\BIBforeignlanguage{en-US}{Dermaknet: Incorporating the knowledge of dermatologists to convolutional neural networks for skin lesion diagnosis},'' \emph{\BIBforeignlanguage{en-US}{IEEE J. Biomed. Health Inform.}}, p. 547–559, Mar 2019.

\bibitem{ner1_2022}
H.~Chen, W.~Zhang, L.~Cheng, and H.~Ye, ``Diverse and high-quality data augmentation using gpt for named entity recognition,'' in \emph{Commun. Comput. Info. Sci.}\hskip 1em plus 0.5em minus 0.4em\relax Springer, 2022, pp. 272--283.

\bibitem{ner2_2023}
S.~Wang, X.~Sun, X.~Li, R.~Ouyang, F.~Wu, T.~Zhang, J.~Li, and G.~Wang, ``Gpt-ner: Named entity recognition via large language models,'' \emph{arXiv}, 2023.

\bibitem{ouyang2022instructgpt}
L.~Ouyang, J.~Wu, X.~Jiang, D.~Almeida, C.~Wainwright, P.~Mishkin, C.~Zhang, S.~Agarwal, K.~Slama, A.~Ray \emph{et~al.}, ``Training language models to follow instructions with human feedback,'' \emph{Adv. neural inf. proces. syst.}, vol.~35, pp. 27\,730--27\,744, 2022.

\bibitem{li2022grounded}
L.~H. Li, P.~Zhang, H.~Zhang, J.~Yang, C.~Li, Y.~Zhong, L.~Wang, L.~Yuan, L.~Zhang, J.-N. Hwang \emph{et~al.}, ``Grounded language-image pre-training,'' in \emph{Proc. IEEE Conf. Comput. Vis. Pattern Recognit.}, 2022, pp. 10\,965--10\,975.

\bibitem{qin2022medical}
Z.~Qin, H.~H. Yi, Q.~Lao, and K.~Li, ``Medical image understanding with pretrained vision language models: A comprehensive study,'' in \emph{Proc. Int. Conf. Learn. Represent.}, 2023, pp. 1--13.

\bibitem{chen2023knowledge}
X.~Chen, Y.~He, C.~Xue, R.~Ge, S.~Li, and G.~Yang, ``Knowledge boosting: Rethinking medical contrastive vision-language pre-training,'' in \emph{Lect. Notes Comput. Sci.}\hskip 1em plus 0.5em minus 0.4em\relax Springer, 2023, pp. 405--415.

\bibitem{shih2019augmenting}
G.~Shih, C.~C. Wu, S.~S. Halabi, M.~D. Kohli, L.~M. Prevedello, T.~S. Cook, A.~Sharma, J.~K. Amorosa, V.~Arteaga, M.~Galperin-Aizenberg \emph{et~al.}, ``Augmenting the national institutes of health chest radiograph dataset with expert annotations of possible pneumonia,'' \emph{Radiology: Artificial Intelligence}, vol.~1, no.~1, p. e180041, 2019.

\bibitem{wang2017chestx}
X.~Wang, Y.~Peng, L.~Lu, Z.~Lu, M.~Bagheri, and R.~M. Summers, ``Chestx-ray8: Hospital-scale chest x-ray database and benchmarks on weakly-supervised classification and localization of common thorax diseases,'' in \emph{Proc. IEEE Conf. Comput. Vis. Pattern Recognit.}, 2017, pp. 2097--2106.

\bibitem{nguyen2022vindr}
H.~Q. Nguyen, K.~Lam, L.~T. Le, H.~H. Pham, D.~Q. Tran, D.~B. Nguyen, D.~D. Le, C.~M. Pham, H.~T. Tong, D.~H. Dinh \emph{et~al.}, ``Vindr-cxr: An open dataset of chest x-rays with radiologist’s annotations,'' \emph{Sci. Data}, vol.~9, no.~1, p. 429, 2022.

\bibitem{rahman2021exploring}
T.~Rahman, A.~Khandakar, Y.~Qiblawey, A.~Tahir, S.~Kiranyaz, S.~B.~A. Kashem, M.~T. Islam, S.~Al~Maadeed, S.~M. Zughaier, M.~S. Khan \emph{et~al.}, ``Exploring the effect of image enhancement techniques on covid-19 detection using chest x-ray images,'' \emph{Comput. Biol. Med.}, vol. 132, 2021.

\bibitem{goldberger2000physiobank}
A.~L. Goldberger, L.~A. Amaral, L.~Glass, J.~M. Hausdorff, P.~C. Ivanov, R.~G. Mark, J.~E. Mietus, G.~B. Moody, C.-K. Peng, and H.~E. Stanley, ``Physiobank, physiotoolkit, and physionet: components of a new research resource for complex physiologic signals,'' \emph{Circ.}, vol. 101, no.~23, pp. e215--e220, 2000.

\bibitem{resnet}
K.~He, X.~Zhang, S.~Ren, and J.~Sun, ``\BIBforeignlanguage{en-US}{Deep residual learning for image recognition},'' in \emph{\BIBforeignlanguage{en-US}{Proc. IEEE Conf. Comput. Vis. Pattern Recognit.}}, Jun 2016, pp. 770--778.

\bibitem{convnext}
Z.~Liu, H.~Mao, C.-Y. Wu, C.~Feichtenhofer, T.~Darrell, and S.~Xie, ``A convnet for the 2020s,'' in \emph{Proc. IEEE Conf. Comput. Vis. Pattern Recognit.}, 2022, pp. 11\,976--11\,986.

\bibitem{biobert_2019}
E.~Alsentzer, J.~Murphy, W.~Boag, W.-H. Weng, D.~Jindi, T.~Naumann, and M.~McDermott, ``Publicly available clinical bert embeddings,'' in \emph{Proc. Annu. Meet. Assoc. Comput Linguist.}, 2019, pp. 72--78.

\bibitem{transformer}
A.~Vaswani, N.~Shazeer, N.~Parmar, J.~Uszkoreit, L.~Jones, A.~Gomez, L.~Kaiser, and I.~Polosukhin, ``\BIBforeignlanguage{en-US}{Attention is all you need},'' in \emph{\BIBforeignlanguage{en-US}{Proc. Mach. Learn. Res.}}, Jun 2017.

\bibitem{pneumothorax}
\BIBentryALTinterwordspacing
``Siim-acr pneumothorax segmentation,'' 2019. [Online]. Available: \url{https://www.kaggle.com/c/siim-acr-pneumothorax-segmentation}
\BIBentrySTDinterwordspacing

\end{thebibliography}

\end{document}